\documentclass{article}
\usepackage{spconf,amsmath,graphicx}
\usepackage{verbatim}
\usepackage{comment}
\usepackage{hyperref}
\usepackage{amsfonts}
\usepackage{footnote}

\title{Hand-object reconstruction via \linebreak interaction-aware graph attention mechanism}
\name{Taeyun Woo$^1$ \qquad Tae-Kyun Kim$^{1,2}$ \qquad Jinah Park$^{1}$}
  
\address{$^1$School of Computing, KAIST, Republic of Korea
\\ $^2$Imperial College London, UK}
\begin{document}
%
\maketitle

\let\thefootnote\relax\footnotetext{Email: \{taeyun.woo, kimtaekyun, jinahpark\}@kaist.ac.kr}
\let\thefootnote\relax\footnotetext{Acknowledgements: This work was supported by the National Research Council of Science \& Technology (NST) grant by the Korea government (MSIT) (CRC23021-000), NST grant (CRC 21011, MSIT), KOCCA grant (R2022020028, MCST), and IITP grant (RS-2023-00228996, MSIT).}

\begin{abstract}
Estimating the poses of both a hand and an object has become an important area of research due to the growing need for advanced vision computing. The primary challenge involves understanding and reconstructing how hands and objects interact, such as contact and physical plausibility.
Existing approaches often adopt a graph neural network to incorporate spatial information of hand and object meshes. However, these approaches have not fully exploited the potential of graphs without modification of edges within and between hand- and object-graphs. 
We propose a graph-based refinement method that incorporates an interaction-aware graph-attention mechanism to account for hand-object interactions. Using edges, we establish connections among closely correlated nodes, both within individual graphs and across different graphs. Experiments demonstrate the effectiveness of our proposed method with notable improvements in the realm of physical plausibility.

\end{abstract}
\section{Introduction}
\label{sec:intro}

Hand-object pose estimation is a promising research field for VR/AR \cite{VR1}, human-computer interaction \cite{HCI1}, and robotics \cite{R1}. 
The primary goal is to understand and make the hand interact with its surrounding objects.
Previous approaches \cite{HO1, hasson, HO2, FPHA, HO3, HO4, colab, coarse, wacv} estimated hand and object poses from images under hand-object interaction (HOI) scenarios. 
At an early stage, these poses are estimated as a skeletal chain by connecting hand joints. However, it is too sparse to represent HOI, such as contact regions on the hand and the object's surface. For this reason, recent methods \cite{hasson, HO2, HO3, HO4, colab, weakly} predicted 3D shapes that convey dense surface points. 

In the context of HOI reconstruction, such as contacts and physical plausibility, the representation should be high-dimensional \textit{i.e.} dense surface points or meshes, since HOI occurs on the surfaces of the instances. In addition, physically feasible hand-object shapes and poses are essential to improve usability and stability for practical applications.
Several approaches \cite{HO2, HO4, colab, coarse, wacv} have incorporated a feature fusion scheme that combines the latent features of the hand and the object. This scheme enables accurate estimation of hand and object poses, considering their interaction through jointly combined features. Furthermore, it can improve physical plausibility, \textit{e.g.} minimizing penetration volume. Typically, there are two methods for feature fusion: (1) latent feature fusion and (2) graph-based feature fusion.

The latent feature fusion schemes \cite{HO2, HO4, colab} utilize two separate encoders to extract each latent feature of the hand and object from an input image. Subsequently, two features are fused, and hand- and object-poses are estimated from the combined feature. This technique improves physical plausibility compared to those estimating the hand and object independently. However, it is hard to explicitly control the fusion of distinct features within the network.

To address this limitation, some methods \cite{colab, coarse, wacv} adopted graph neural networks (GNNs) \cite{GNN,GCN}. This technique is advantageous when dealing with data structures that involve spatial information and connectivity, such as meshes. 
Although graph-based methods show improvements over latent feature fusion, the fusion of hand and object features typically occurs during the graph initialization \cite{colab} or node initialization stages \cite{wacv}, limiting the consideration of node connectivity. An alternative approach by Zhang et al. \cite{coarse} modified the hand and object graph by adding new edges. However, their method is limited to estimating \textit{sparse} hand-object poses: hand joints and the object bounding box.

Previous graph-based approaches focused on incorporating appropriate features into the mesh-structured graph nodes of the hand graph and object graph, neglecting the consideration of node connectivity. On the other hand, our approach aims to enhance physical plausibility by introducing node-connecting schemes that link highly correlated nodes. We define two types of edges: common relation edges $E_c$ and attention-guided edges $E_a$. These edges link not only intra-class (each hand or object separately) nodes but also inter-class (hand-object) nodes. This distinguishes our method from previous approaches \cite{colab}, which focused solely on intra-class edges. Our interaction-aware graph mechanism considers hand-object interactions, improving the physical plausibility between a hand and an object.

We have evaluated our approach on ObMan \cite{hasson} and DexYCB \cite{dexycb}. The results show improved physical plausibility between the hand and object. Our contribution is summarized as follows:
    1) We propose an interaction-aware graph attention mechanism to enhance physical plausibility for hand-object pose estimation.
    2) Our graph refinement considers the hand-object interaction with common relation edges and attention-guided edges between both intra-class nodes and inter-class nodes.
    3) Finally, we show that our method improves physical plausibility with quantitative and qualitative results.

\begin{figure*}[t]
    \centering
    \centerline{\includegraphics[width=16.5cm]{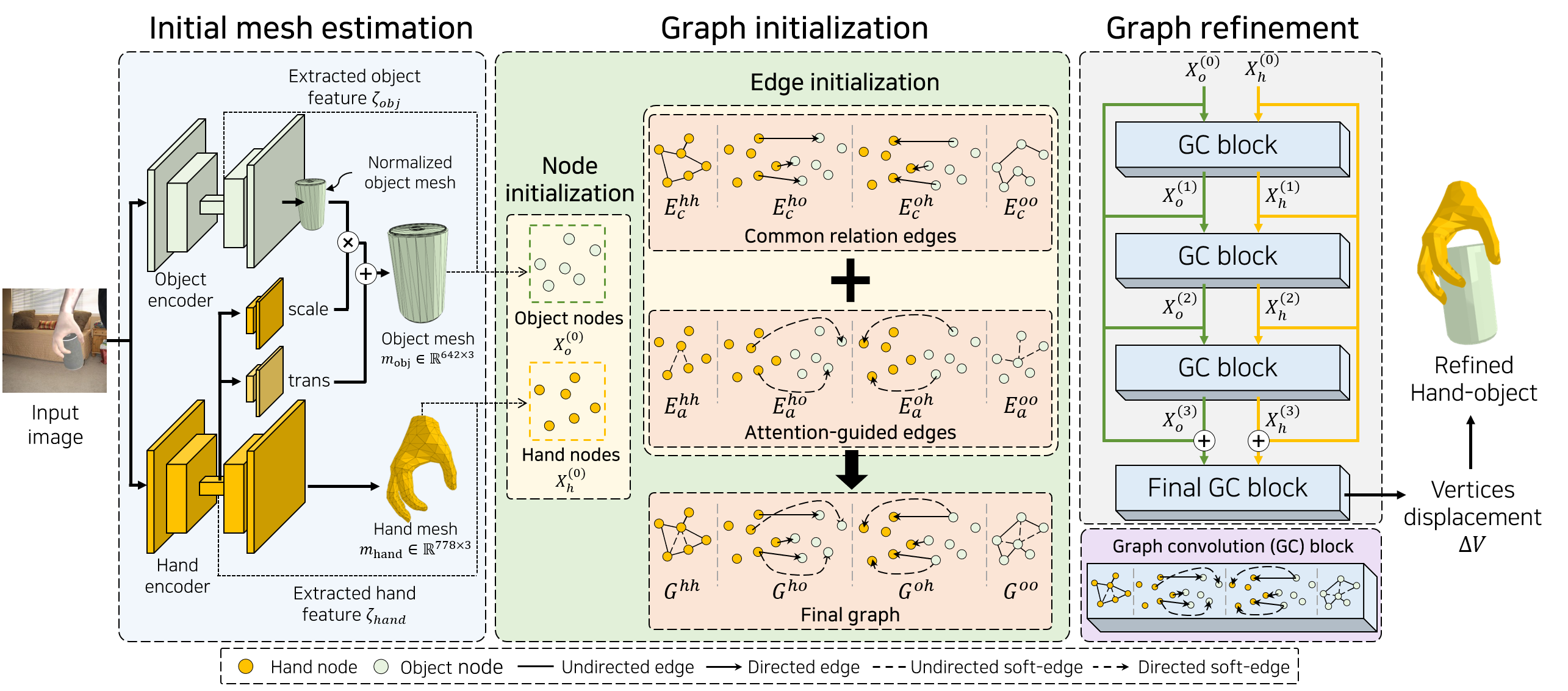}}
    \vspace{-0.2cm}
    \caption{
        Overview of the proposed method.
    }\medskip
    \vspace{-0.3cm}
    \label{fig:network}
\end{figure*}

\section{Related works}
\label{sec:RW}


In earlier days, hand and object poses were predicted in the form of hand joints and a 3D bounding box of the object. However, these representations are too sparse to adequately illustrate the hand-object interaction, since they do not fully capture the hand-object shape. Instead of estimating sparse poses, numerous approaches \cite{hasson, HO2, HO3, HO4, colab} have emerged to estimate hand-object 3D shapes. Hasson et al. \cite{hasson} first suggested both hand- and object-mesh estimation techniques. They estimated the hand mesh with MANO \cite{mano}, a parameterized hand model; and the object mesh is acquired through AtlasNet \cite{atlas}. We adopted their framework to predict the initial hand and object meshes. 

Previous methods estimate the hand and object meshes in a separate branch independently. However, these methods are inherently limited to the consideration of HOI. The hand and object themselves give key insights into each other: the hand pose is highly correlated with the object's shape and features, and vice versa. In this respect,  \cite{HO2, HO3, colab, wacv} proposed a feature fusion technique between a hand and an object where each hand and object is estimated from the combined feature of the hand and object. 

\noindent \textbf{Latent feature fusion.} 
Chen et al. \cite{HO4} reconstructed hand-object meshes from an image using latent feature fusion on top of the network architecture of Hasson et al. \cite{hasson}. In contrast to Hasson et al., Chen et al.'s network estimated hand and object meshes by fusing the features of hand and object through an LSTM module. This fusion technique reduces the inter-penetration volume between the hand and the object. Liu et al. \cite{HO2} introduced the attention \cite{att} based feature fusion scheme to improve the feature of an object by the feature of a hand. In this work, we consider hand-object interaction, connecting highly correlated nodes to incorporate each feature of the hand and object. We employ a graph-based feature fusion rather than the latent feature fusion scheme.

\noindent \textbf{Graph-based feature fusion.}
Graph-based feature fusion \cite{HO1, colab, coarse, wacv} combines hand and object features using a GNN on a graph structure, while the latent feature fusion scheme integrates features in the latent space. Most methods in this domain estimate initial hand and object meshes (or joints), and construct graphs whose structures are the same as the estimated individual meshes. Given the two sub-graph structures, the feature fusion is done by injecting both features to each node of graphs. Subsequently, GNN layers operate on these graphs, and it outputs the feature vector which is used to refine the initially estimated hand object meshes. For instance, Tse et al. \cite{colab} refined the initial meshes with the image feature and the graph feature generated by the GNN layers. Wang et al. \cite{wacv} constructed the hand and object graphs, and refined the initial meshes through cross-attention between the hand and object graph nodes.

Most methods, however, maintain a static graph structure, solely relying on the initially estimated hand and object meshes without structural changes. Though Tse et al. have used a different graph from the initial meshes by newly connecting the nodes, it only involves the modification between intra-class nodes (i.e. within each hand or object), not inter-class nodes. While Zhang et al. \cite{coarse} proposed edge modifications connecting both intra-class and inter-class nodes, they estimated hand joints and an object bounding box, limiting their understanding of hand-object interactions. In contrast, our method incorporates not only mesh-structured edges but also additional edges between intra-class nodes and inter-class nodes. We demonstrate the improvement in physical plausibility between the hand and the object via our interaction-aware graph-attention mechanism.  

\section{INTERACTION-AWARE GRAPH ATTENTION}
\label{sec:method}
The proposed method estimates hand and object meshes from a single image, and fuses both the hand and object features with an interaction-aware graph attention mechanism.
Our network architecture is shown in Fig. \ref{fig:network}. In the following, we describe each part of our method in detail.


\subsection{Initial mesh estimation}
The hand mesh and object mesh are predicted in independent branches, and predicted meshes are on a hand-wrist coordinate, where the root of the hand (wrist) is the origin. 
Note that we follow Hasson et al.'s \cite{hasson} architecture design to estimate the initial meshes. 

\noindent \textbf{Hand mesh estimation.} From an input image, ResNet-18 \cite{resnet} extracts a hand feature vector $\zeta_{\text{hand}}$. The hand feature is fed into the MANO \cite{mano} differential layer \cite{hasson}, and outputs the MANO pose and shape parameters. The hand mesh $m_{\text{hand}}$ is reconstructed from these parameters.  

\noindent \textbf{Object mesh estimation.} As in Hasson et al. \cite{hasson}, we first pre-train the AtlasNet \cite{atlas} with normalized point clouds. ResNet-18 extracts an object feature vector $\zeta_{\text{obj}}$ and the feature is fed into AtlasNet. Then, AtlasNet estimates the normalized object mesh $m_{\text{obj}}^n$. Since all object meshes have different numbers of vertices, we prepared uniformly sampled points on the normalized object meshes as the ground truth point cloud. Finally, the scale and translation are adjusted to the normalized object mesh, estimated from the hand feature $\zeta_{\text{hand}}$.


\subsection{Graph initialization}
Graph initialization is two-fold: node initialization and edge initialization. These are then utilized in the graph refinement stage.

\noindent \textbf{Notations.} A graph $G$ is defined as $G=(N, E)$ where $N$ is a set of nodes $n$, and $E$ is a set of edges $e$. Let us define the hand graph $G_{\text{hand}}=(N_{\text{hand}}, E_{\text{hand}})$ and the object graph $G_{\text{obj}}=(N_{\text{obj}}, E_{\text{obj}})$. Within the graphs, each node is represented by $n_{\text{hand}}^i \in N_{\text{hand}}$ and $n_{\text{obj}}^i \in N_{\text{obj}}$, where $i$ denotes the node index.

\noindent \textbf{Node initialization.}
To each node, the corresponding vertex coordinate $v$ of the estimated mesh $m$ and the extracted image feature $\zeta$ are injected. Especially, for each node $i$ of hand and object graphs, the hand node feature $x_{\text{hand}}^i$ and object node feature $x_{\text{obj}}^i$ are defined as follows:
$ x_{\text{c}}^i \leftarrow  v_{\text{c}}^{i} \oplus \zeta_{\text{c}}; \; c\in {\{\text{hand, obj}\}}, $
where $\leftarrow$ refers to the injection of features and $\oplus$ represents the concatenation. 
The nodes' feature sets are denoted as $X_{\text{hand}}$ and $X_{\text{obj}}$ for the hand and the object.

\noindent \textbf{Edge initialization.}
We connect nodes to consider hand-object interactions with two types of edges: common relation edges $E_c$ and attention-guided edges $E_a$. Both types of edges link nodes not only within the same class (hand-hand and object-object) but also across different classes (hand-object and object-hand). $E_c$ is constructed by the spatial distance between the graph nodes, representing common adjacent behaviors. However, relying solely on common relation edges may miss contextual or global features of hand-object interactions. Therefore, we introduce $E_a$ based on the distance in a latent space defined by the attention mechanism \cite{att}.

For both $E_c$ and $E_a$, there are four types of edges: $E^{hh}$, $E^{oo}$, $E^{ho}$, and $E^{oh}$. The upper subscript denotes the type of node, where $h$ represents a hand and $o$ represents an object; $ho$ indicates edges from a hand node to an object node. Additionally, $E^{hh}$ and $E^{oo}$ are homogeneous undirected edges, while $E_c^{ho}$ and $E_c^{oh}$ are heterogeneous directed edges. During the graph refinement stage, the combination of these two edge sets (from $E_a$ and $E_c$) is considered as the final graph's edge set.

\noindent \textbf{Common relation edges $E_{c}$.}
There are four kinds of common relation edges: $E_c^{hh}$, $E_c^{oo}$, $E_c^{ho}$, and $E_c^{oh}$. The edge sets $E_{c}^{hh}$ and $E_{c}^{oo}$ represent the mesh faces $f_{hand}$ and $f_{obj}$, respectively.

In contrast, $E_c^{ho}$ and $E_c^{oh}$ are initialized, connecting the edges between the nearest node pairs from hand to object nodes, and vice versa. This scheme looks similar with Zhang et al. \cite{coarse}, however, we do not connect all possible edges between every closest pair of nodes each other. There are two reasons behind this design. First, Zhang et al. estimated hand joints and an object bounding box that is a sparse representation than meshes. Hence, if we connect all pairs of dense mesh nodes, it computationally costs. The other reason is that we aim to connect only relevant nodes. Therefore, we selectively connect pairs of nodes based on the contact prior. This strategy ensures computational efficiency and captures meaningful interactions between the hand and object.

For $E_{c}^{ho}$, we connect the edges from designated hand nodes to their nearest object nodes. These designated hand nodes are frequently in contact with an object in ObMan \cite{hasson} dataset. Unlike hands, objects do not have specific contact regions on their surfaces. To address this, we simply reverse the edges of $E_c^{ho}$ as $E_c^{oh}$. We also adopt this strategy to DexYCB \cite{dexycb} training.

\noindent \textbf{Attention-guided edges $E_a$.}
Similarly, There are four attention-guided edges: $E_a^{hh}$, $E_a^{oo}$, $E_a^{ho}$, and $E_a^{oh}$. The attention-guided edges are determined by values of the attention matrix \cite{att}. Specifically, $E_a^{hh}$ and $E_a^{oo}$ are built on self-attention matrices, applying self-attention to node features $X_{\text{hand}}$ and $X_{\text{obj}}$, respectively. In contrast, $E_a^{ho}$ and $E_a^{oh}$ are constructed from cross-attention matrices from hand node features $X_{\text{hand}}$ to object node features $X_{\text{obj}}$, and vice versa. Each value within these attention matrices serves as edge weights. This type of edge is referred to as a \textit{soft} edge, denoting its weight less than 1. At the aggregation stage, the neighbor node features are aggregated by multiplying the edge weight of the connected edges.

Note, not all elements in the attention matrices are used as edges. The attention values are obtained by passing the softmax operator, which ensures that the sum of each row or column of the matrix equals 1. Since the neighbor node features are aggregated, multiplying the corresponding edge weight, there exist negligible values in the matrix. For this reason, we only use the attention values that are greater than the threshold $\gamma$. We empirically set $\gamma$ as $0.01$.

\noindent \textbf{Final graph design.} In summary, we construct four kinds of graphs: $G^{hh},\, G^{ho},\, G^{oh},\, \text{and} \,  G^{oo}$. Each set of graph nodes stands for four kinds of nodes: hand-hand, hand-object, object-hand, and object-object. And their edge sets are the union of corresponding common relation edges $E_c$ and attention-guided edges $E_a$ based on the graph nodes. For instance, the edge set of $G^{ho}$ is $E_c^{ho}\cup E_a^{ho}$. The edge connectivity is integrated with the union operation, while the edge weight is integrated with the addition operation.

\begin{figure}[t]
\begin{minipage}[b]{1\linewidth}
  \centering
  \centerline{\includegraphics[width=0.8\linewidth]{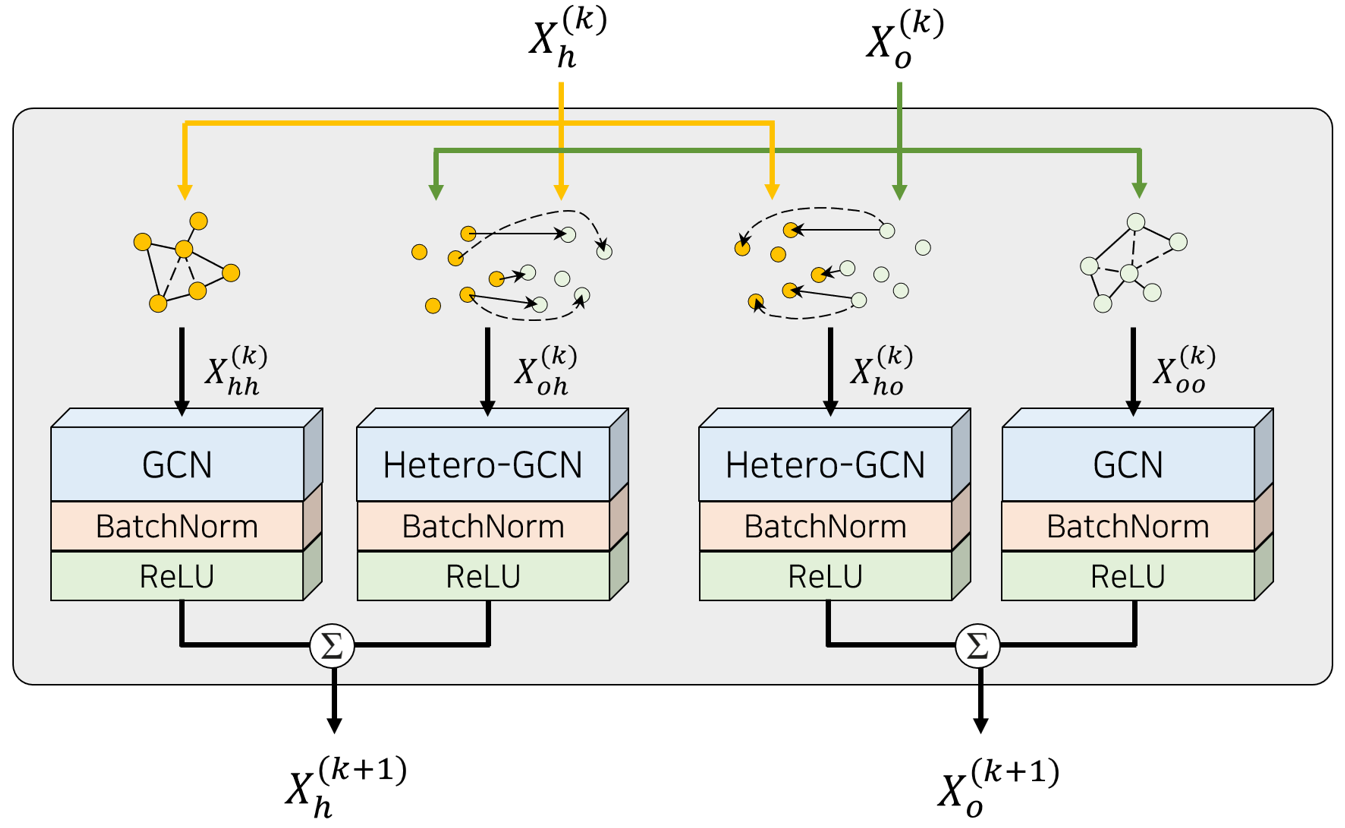}}
  \centerline{(a) GC blocks for $k \in \{1,2,3\}$}\medskip
\end{minipage}
\begin{minipage}[b]{1\linewidth}
  \centering
  \centerline{\includegraphics[width=0.75\linewidth]{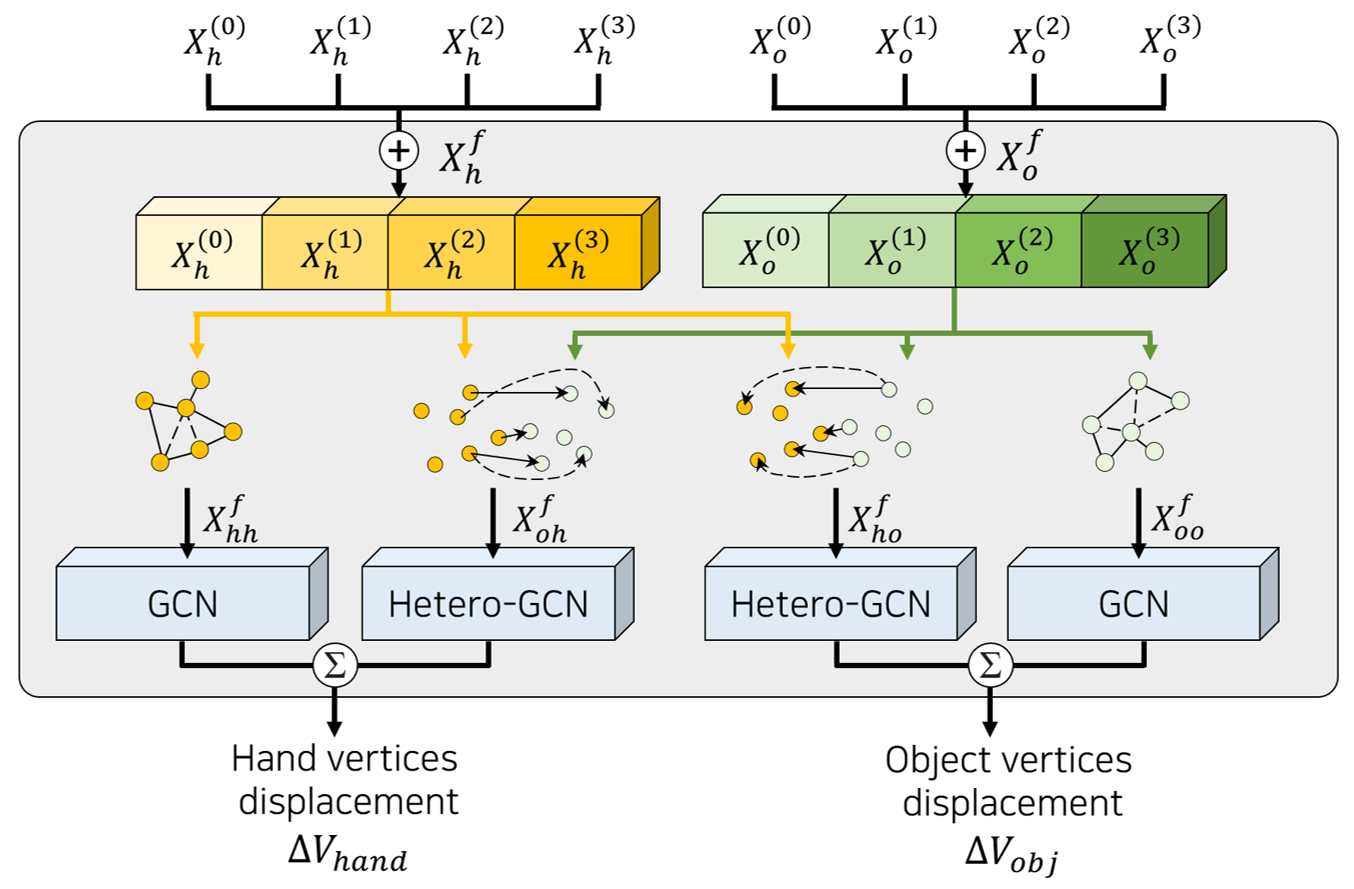}}
  \centerline{(b) Final GC block}\medskip
\end{minipage}
\hfill
\vspace{-0.8cm}
\caption{Design of Graph convolution (GC) blocks.}
\label{fig:gcblock}
\vspace{-0.5cm}

\end{figure}

\subsection{Refinement stage}\label{sec:refinement}
Graph refinement stage gets hand node features $X_{\text{hand}}$ and object node features $X_{\text{obj}}$, and outputs vertex displacements through four graph convolution (GC) blocks. Each GC block consists of four GCN layers, operating on the final graph design ($G^{hh}$, $G^{oh}$, $G^{ho}$, and $G^{oo}$). The last GC block estimates vertex displacements which are added to corresponding vertices of hand mesh and object mesh for the refinement. The structure of GC block is illustrated in Fig. \ref{fig:gcblock}. 

\noindent \textbf{Preliminary.}
The graph convolution \cite{GCN} operates by updating node features through a message passing scheme. This scheme consists of two steps: aggregation and update. The aggregation gathers the node features of one node $i$ and its neighboring nodes $\mathcal{N}(i)$, and encapsulated into a message $msg_{i}^{(k)}$ along node $i$ as follows:
$$ msg_{i}^{(k)} = AGGREGATE^{(k)} (n_{p}^{(k-1)}, p \in (\mathcal{N}(i) \cup \{i\}) ).$$
Here, $n_{p}^{(k-1)}$ represents the feature of node $p$ from the $(k-1)$-th layer, and $\mathcal{N}(i)$ denotes the set of neighboring nodes of node $i$. The aggregate function $AGGREGATE$ combines the features of neighboring nodes by multiplying each node's feature with its corresponding edge weight, and then summing them. Following the aggregation step, the update stage modifies the node feature at the $k$-th layer as follows:

$$ n_{i}^{(k)} = UPDATE^{(k)} (n_{i}^{(k-1)}, msg_{i}^{(k)}). $$
The update function $UPDATE$ changes the node feature to the aggregated message at the current layer. This message passing scheme allows information to flow between neighboring nodes, enabling the graph convolution layer to capture relational dependencies and refine node features accordingly. 

\noindent \textbf{Graph convolution block.}
As in Fig. \ref{fig:network}, we employed four graph convolution blocks, and each layer consists of four GCNs. Each GCN layer receives the corresponding node features and operates on the final graph structure: $G^{hh}$, $G^{oh}$, $G^{ho}$, and $G^{oo}$. Except for the final layer, the other layers get a previous stage's hand node features $X_h^{(k-1)}$ and object node features $X_o^{(k-1)}$, then output updated node features $X_h^{(k)}$ and $X_o^{(k)}$. Whereas, the final layer gets all previous node features from the preceding layers. This design choice mitigates the graph over-smoothing \cite{oversmoothing}, where all nodes in the graph tend to converge to a single value as more graph convolution layers are stacked. As a result, the final GC block outputs vertex displacements, and the displacements are added to the corresponding vertices of hand mesh and object mesh respectively.

\subsection{Training losses}
\noindent \textbf{Hand loss $\mathcal{L}_{\text{hand}}$.} 
The hand loss $\mathcal{L}_{\text{hand}}$ comprises three components: hand mesh vertices loss $\mathcal{L}_{v}$, the L2 term for hand joints loss $ \mathcal{L}_{J}$, and a regularization loss term for shape parameters $\mathcal{L}_{\beta}$. $\mathcal{L}_{v}$ and $\mathcal{L}_{J}$ guide the estimated hand mesh vertices $v_{\text{hand}}$ and hand joint $J_{\text{hand}}$ to be similar to the ground truth values by L2 distance. The regularization loss $\mathcal{L}_{\beta}=||\beta||^2$ incorporates shape parameter $\beta$ to be close to the average shape in the MANO hand set.  The hand reconstruction loss is a summation of these three terms: $ \mathcal{L}_{\text{hand}} = \mathcal{L}_{v} + \mathcal{L}_{J} + \mathcal{L}_{\beta}.$

\noindent \textbf{Object pre-training loss $\mathcal{L}_{m_{\text{obj}}}$.} 
$\mathcal{L}_{m_{\text{obj}}}$ is utilized to pre-train the AtlasNet with normalized point clouds. It consists of the mean symmetric chamfer distance loss $\mathcal{L}_{CD}$, edge regularizing loss $\mathcal{L}_{e}$, and Laplacian smoothing loss $\mathcal{L}_{L}$. $\mathcal{L}_{CD}$ ensures the estimated object mesh aligns with the ground truth. $\mathcal{L}_{e}$ encourages the uniform edge lengths in the object mesh, and $\mathcal{L}_{L}$ guides the reconstructed mesh faces to be smooth, and $\mathcal{L}_{m_{\text{obj}}}$ is the sum of these terms with weights $\lambda_{e}=2$ and $\lambda_{L}=0.1$ as in the previous work \cite{hasson}: $ \mathcal{L}_{m_{\text{obj}}} = \mathcal{L}_{CD} + \lambda_{e}\mathcal{L}_{e} + \lambda_{L}\mathcal{L}_{L}. $

\noindent \textbf{Object training loss $\mathcal{L}_{\text{obj}}$.} 
Unlike pre-training, the final object mesh is defined on the hand-wrist coordinate, hence, we have to consider the coordinate-aligned ground truth. We additionally estimate the scale and translation to modify the normalized mesh. As we adopted Hasson et al.'s \cite{hasson} framework, we followed their loss settings. In summary, the object branch is trained by object training loss 
 $\mathcal{L}_{\text{obj}} = \mathcal{L}_{m_{\text{obj}}} + \mathcal{L}_{\text{trans}} + \mathcal{L}_{\text{scale}}$.

\noindent \textbf{Refinement loss $\mathcal{L}_{\text{refine}}$.}
Refinement loss $\mathcal{L}_{\text{refine}}$ enforces the refined hand mesh and refined object mesh similar with the ground truth. $\mathcal{L}_{\text{refine}}$ is the summation of $\mathcal{L}_{\text{hand}}'$ and $\mathcal{L}_{\text{obj}}'$, while $'$ denotes the loss between the refined pose and the ground truth pose. Unlike $\mathcal{L}_{\text{hand}}$, $\mathcal{L}_{\text{hand}}'$ does not contain $\mathcal{L}_{\beta}$, as we did not utilize MANO during the refinement stage.

\section{Experiments}
We present experimental details on the implementation and datasets. Subsequently, comparisons with the baselines and results of the ablation study are presented.

\subsection{Experimental details}
\noindent \textbf{Implementation details.}
We trained the AtlasNet with normalized object point clouds at a learning rate $10^{-4}$ for 200 epochs. During ObMan dataset training, the network is trained at a learning rate $10^{-4}$ for 200 epochs, and it decreases to $10^{-5}$ until the next 100 epochs are finished. During DexYCB dataset training, the network is trained at a learning rate $10^{-4}$ for 20 epochs. We employed the Adam optimizer \cite{adam} to all experiments. When we trained AtlasNet alone, we supervised the normalized object estimation with $\mathcal{L}_{m_{\text{obj}}}$. The final network is supervised with $\mathcal{L}_{\text{obj}} + \mathcal{L}_{\text{hand}} + \mathcal{L}_{\text{refine}}$. 

\noindent \textbf{Datasets.} We employed two datasets for training and evaluation: ObMan \cite{hasson} and DexYCB \cite{dexycb}. ObMan is a synthetic dataset that illustrates when a hand grasps an object. Our model is trained with 141K images training split of ObMan, and evaluated on 6.2K images in the test split. DexYCB is a real-world dataset that is captured when the hand grasps an object. We used the official split 's0' to train and evaluate our model. As the baseline \cite{colab}, we filtered DexYCB images in which the hand and object are 1$cm$ apart, naming as $\text{DexYCB}^{-}$. There are 300K training images in $\text{DexYCB}^{-}$.

\noindent \textbf{Evaluation metrics.} Our method is evaluated on four metrics with baselines: hand error (H.E.), object error (O.E.), maximum penetration (M.P.), and intersection volume (I.V.). Hand error measures the average error between ground truth hand joints and estimated hand joints. The mean symmetric Chamfer distance (mm) between ground truth object mesh and estimated object mesh is used as object error. For physical plausibility, we adopt maximum penetration (mm) and intersection volume ($cm^3$). The maximum penetration measures the maximum penetrated depth from hand mesh vertices to the object mesh's surface when a collision occurs. If not, this value becomes 0. The intersection volume measures intersected volume between the voxelized hand and object with a voxel size of $0.5cm$. Additionally, we employ hand mesh error which measures the mean error between the hand meshes for self-evaluation.

\begin{figure}[htb!]
    \centering
    \centerline{\includegraphics[width=0.95\linewidth]{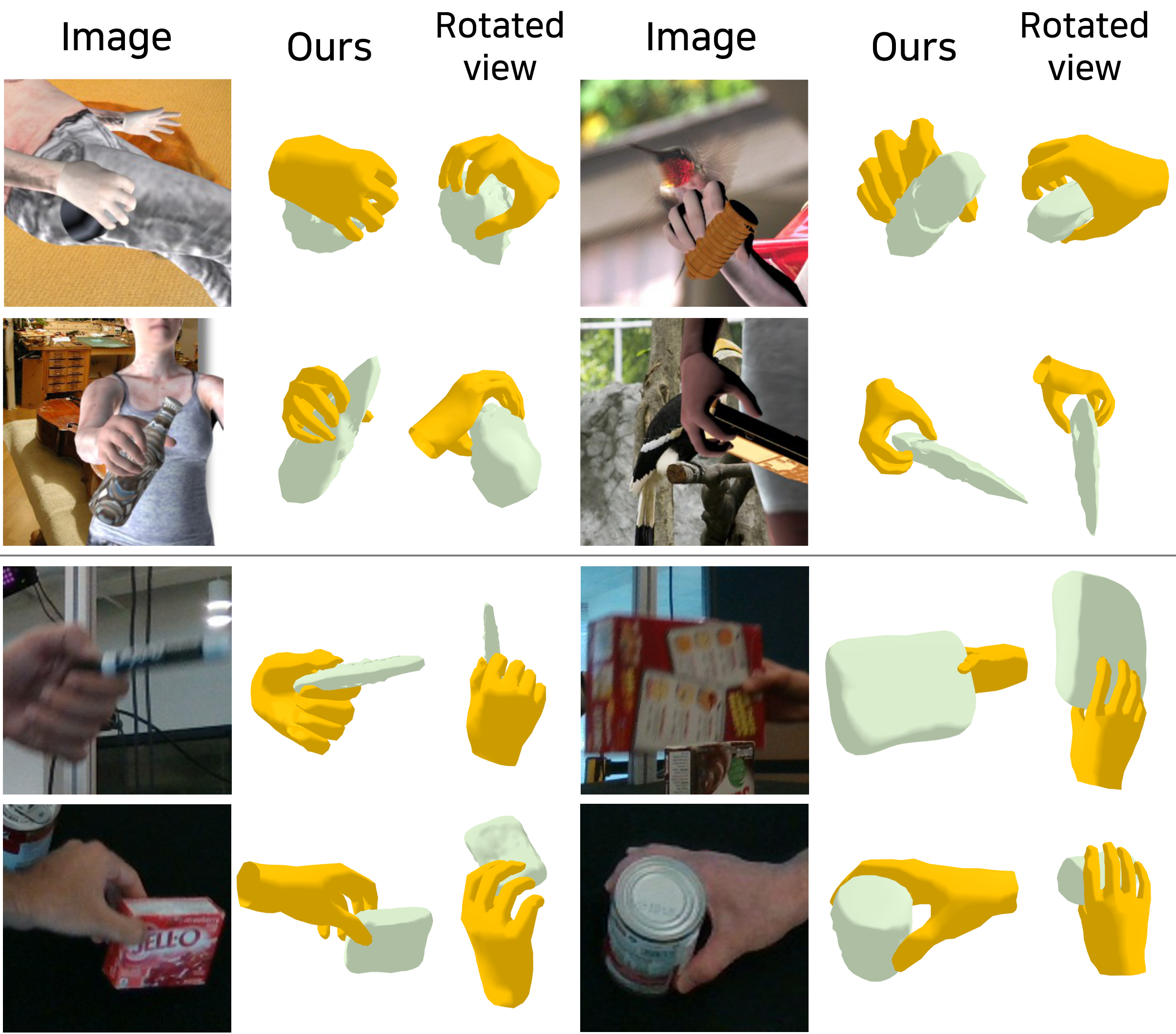}}
    \vspace{-0.2cm}
    \caption{
        Qualitative results of ObMan \cite{hasson} (1-2 rows) and DexYCB \cite{dexycb} (3-4 rows).
    }\medskip
    \vspace{-0.2cm}
    \label{fig:main_result}
    \vspace{-0.2cm}
    
\end{figure}

\begin{figure}[htb!]
    \centering
    \centerline{\includegraphics[width=0.95\linewidth]{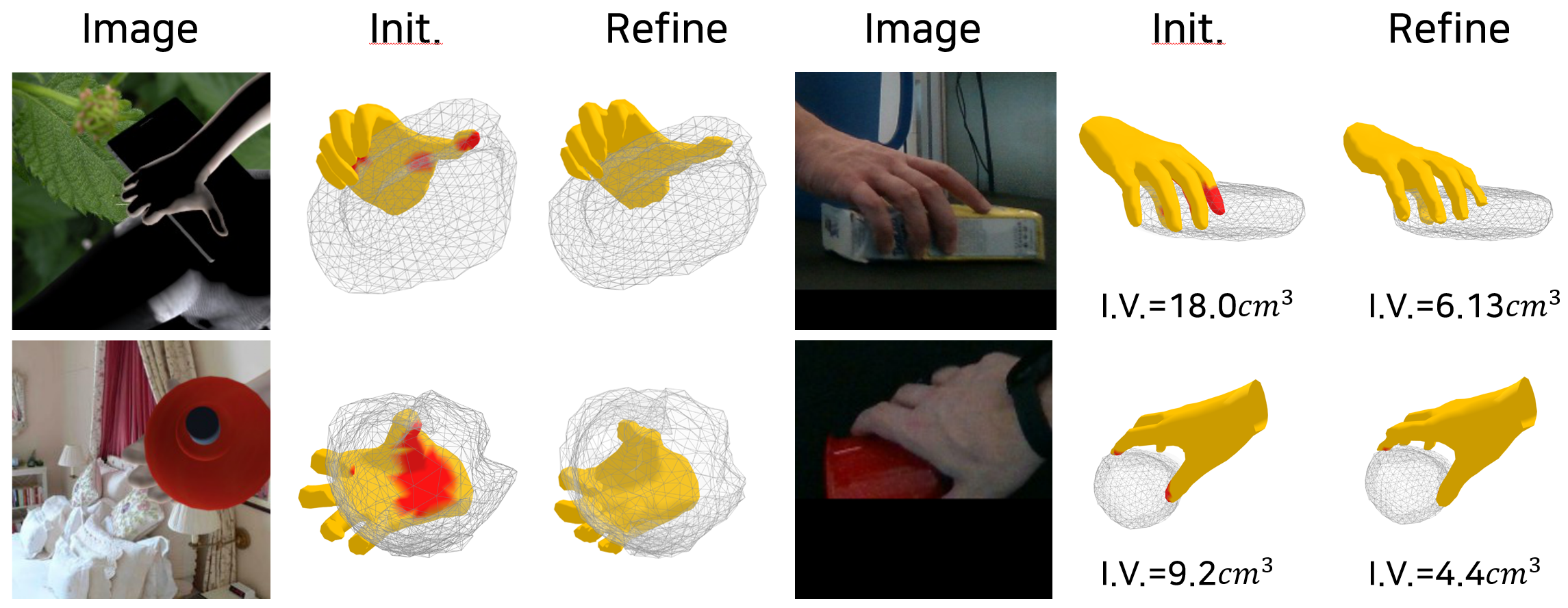}}
    \vspace{-0.2cm}
    \caption{
        Effect of our refinement stage. Red-colored volume represents the intersection volume.
    }\medskip
    \vspace{-0.2cm}
    \label{fig:vispen}
    \vspace{-0.3cm}
    
\end{figure}

\begin{figure}[htb!]
    \centering
    \centerline{\includegraphics[width=0.80\linewidth]{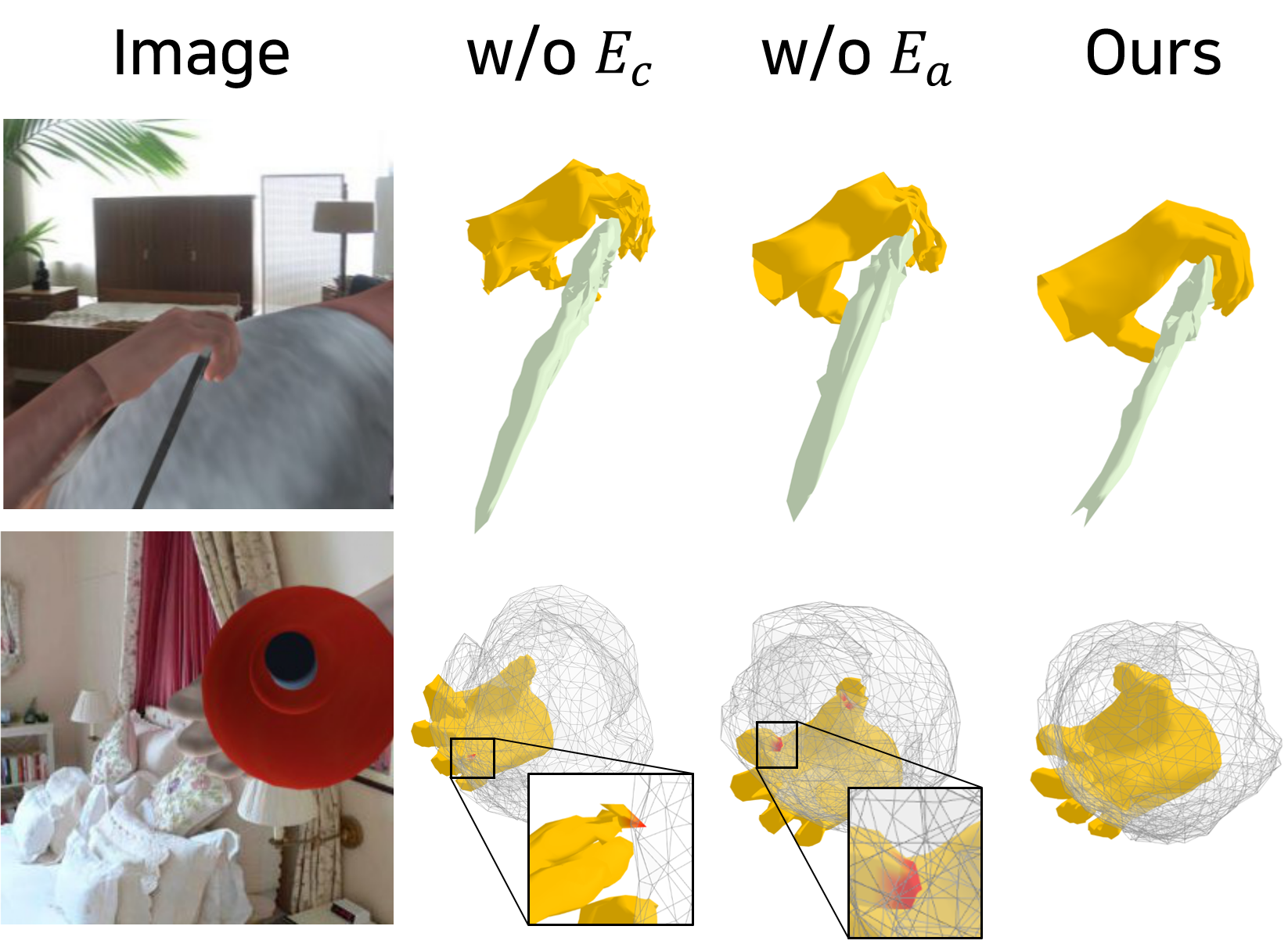}}
    \vspace{-0.2cm}
    \caption{
        Qualitative comparisons of ablation study.
    }\medskip
    \vspace{-0.4cm}
    \label{fig:ablation}
    \vspace{-0.3cm}
    
\end{figure}

\vspace{-0.2cm}
\subsection{Main results}
Our quantitative and qualitative results are presented in Table \hyperref[tab:main_comparison]{1} and Fig. \hyperref[fig:main_result]{3}, respectively. Significantly, our approach surpasses existing baseline methods \cite{hasson, colab} in terms of maximum penetration depth and intersection volume metrics, while our method achieves better than Hasson et al. \cite{hasson} across all evaluation metrics. 

Regarding the hand error, ours is comparable to Tse et al. \cite{colab}, yet the hand error in the ObMan dataset degenerates slightly. This limitation stems from our vertex displacements breaking the MANO hand mesh structure, leading to significant deviations in the positions of the hand joints, particularly at the root position. 
There is a gap between the differences of the object error in ObMan \cite{hasson} and DexYCB \cite{dexycb}, compared to Tse et al. Since ObMan contains 2.7K distinct object meshes, whereas, DexYCB dataset contains fewer available object meshes (20 object meshes) and a scarcity of unique scenarios. For this reason, our model can learn the objects' shape better with ObMan than DexYCB.

Moreover, the annotated hand meshes and object meshes in ObMan do not intersect each other, as ObMan is generated by a physical simulator. However, annotations of DexYCB contain an intersection error between hand meshes and object meshes (M.P.: 4cm / I.V.: 4.3$cm^3$), since DexYCB is collected in the real-world. Hence, our method struggles to learn the hand-object interaction from the DexYCB, as our graph refinement strategy can capture the correlation between a hand and an object.

As mentioned above, ensuring physical plausibility is important for practical applications \cite{VR1, HCI1, R1}, as the instability by the physical implausibility between the hand and object harms the usability than the decline of usability by less accurate hand-object pose. Hence, our method can be more effective than the baselines \cite{hasson, colab} for practical applications.

Furthermore, we evaluate the effectiveness of our refinement stage by comparing between initial estimates and their refined meshes, as illustrated in Table \hyperref[tab:effect_refinement]{2} and Fig. \hyperref[fig:vispen]{4}. Notably, there is a significant improvement in penetration after the refinement stage.

\subsection{Ablation study}
To prove the effect of the common relation edges $E_c$ and the attention-guided edges $E_a$, we trained our network without $E_c$ and $E_a$, respectively. The quantitative and qualitative results are given in Table \hyperref[tab:ablation_study]{3} and Fig. \hyperref[fig:ablation]{5}, respectively. In some metrics, using one of the edges achieves higher results, such as hand joint error in w/o $E_c$. However, when we see the qualitative results in Fig. \hyperref[fig:ablation]{5}, the refined hand-object poses without $E_c$ and $E_a$ are not visually favorable.


\begin{table}[t!]
\label{tab:main_comparison}
\caption{Quantitative comparison with two baselines \cite{hasson, colab} on ObMan \cite{hasson} and DexYCB$^-$ \cite{dexycb} datasets. The baseline results are quoated from Tse et al. \cite{colab}.}
\centering
\resizebox{.49\textwidth}{!}{
\begin{tabular}{l|cccc|cccc}
\hline
\multicolumn{1}{l|}{Datasets}              & \multicolumn{4}{c|}{ObMan} & \multicolumn{4}{c}{DexYCB$^-$} \\ 
\multicolumn{1}{l|}{Metrics}               & H.E.   & O.E.    & M.P.   & I.V.   & H.E. & O.E.   & M.P.     & I.V.     \\ \hline

Hasson et al. \cite{hasson}   &  11.6    & 637.9    & 9.2    &  12.2   &  17.6  & 549.4     &   14.6     &   6.7       \\

Tse et al. \cite{colab}       &  9.1    &  385.7    &  7.4   &  9.3     &  15.3  & 501.2     &  12.1      &    14.9      \\

Ours &  11.2   &  396.1   & 6.9   &  3.5   &15.7        &   534.5  &   9.19  &   5.9       \\ \hline

\end{tabular}
}
\vspace{-0.3cm}
\end{table}

\begin{table}[t]
\label{tab:effect_refinement}
\caption{Effect of our graph refinement. \textit{initial} represents the result with initially estimated meshes and \textit{refine} represents the result with refined hand and object meshes.}
\centering
\small 
\begin{tabular}{l|cc|cc}
\hline
\multicolumn{1}{l|}{Datasets}      & \multicolumn{2}{c|}{ObMan} & \multicolumn{2}{c}{DexYCB$^-$ } \\ 
\multicolumn{1}{l|}{Metrics} & \multicolumn{1}{c}{initial} & \multicolumn{1}{c|}{refine} & \multicolumn{1}{c}{initial} & \multicolumn{1}{c}{refine} \\ \hline
Hand joint error  &    11.0           &      11.2         &     16.7          &    15.8          \\
Hand mesh error  &     13.05           &      10.5         &     17.0          &     15.4         \\
Object error     &     430.1          &       396.1        &     539.6          &   534.5           \\
Max. pen.        &     8.83          &        6.90       &     9.94          &  9.19            \\
Inter. vol.     &      7.72         &        3.51       &     6.4          &  5.9            \\ \hline
\end{tabular}
\vspace{-0.3cm}
\end{table}

\begin{table}[t!]
\label{tab:ablation_study}
\caption{Quantitative results of ablation study on common relation edges $E_c$ and attention-guided edges $E_a$. }

\centering
\resizebox{.48\textwidth}{!}{
\begin{tabular}{l|ccc|ccc}
\hline
                      & \multicolumn{3}{c|}{ObMan} & \multicolumn{3}{c}{DexYCB$^-$ } \\
Metrics & \multicolumn{1}{c}{w/o $E_c$} & \multicolumn{1}{c}{w/o $E_a$} & \multicolumn{1}{c|}{Ours} & \multicolumn{1}{c}{w/o $E_c$} & \multicolumn{1}{c}{w/o $E_a$} & \multicolumn{1}{c}{Ours} \\ \hline
Hand joint error  &   11.2      &   11.1      &  11.2      &   15.6      &  15.5     &   15.7     \\
Hand mesh error  &   11.8      &   10.7      &  10.5      &  15.4      &  15.2     &  15.4      \\
Object error      &   411.5      &    413.4     &   396.1     &  548.7      &  550.8    &   534.5     \\
Max. pen.        &   6.7      &   7.0      &   6.9     &  9.5     &  10.1    &  9.2      \\
Inter. vol.     &   3.2      &   3.6      &   3.5     &  5.1   &  5.8   &   5.9    \\ \hline
\end{tabular}
}
\vspace{-0.3cm}
\end{table}

\section{Conclusion}
We proposed an interaction-aware graph attention mechanism aimed at improving the physical plausibility of hand-object pose estimation. This mechanism incorporates common relation edges and attention-guided edges, fostering connections within intra-class nodes and between inter-class nodes. We have shown that our approach enhances the physical plausibility of estimated hand and object meshes, particularly in addressing issues related to penetration and intersection between them. By improving physical plausibility of the estimated hand and object, both usability and stability are guaranteed for practical applications that need realistic hand-object interaction, such as a hand simulator and imitation learning.

\bibliographystyle{IEEEbib}
\bibliography{refs}

\end{document}